\documentclass[a4paper]{article}
\usepackage{verbatim}
\usepackage{INTERSPEECH_v2}
\newcommand{\fsz}[3]{#1 $\times$ #2 $\times$ #3}
\usepackage{float}
\usepackage[linesnumbered,ruled,vlined]{algorithm2e}
\usepackage{subfig}
\usepackage{makecell}
\usepackage{cite}

\title{A low latency ASR-free end to end spoken language understanding system}

\name{Mohamed Mhiri, Samuel Myer, Vikrant Singh Tomar}
\address{
  Fluent.ai Inc., Montr\'eal, Qu\'ebec, Canada}
\email{mohamed.mhiri@fluent.ai,sam.myer@fluent.ai,vikrant@fluent.ai}

\begin{document}

\maketitle
\begin{abstract}
In recent years, developing a speech understanding system that classifies a waveform to structured data, such as intents and slots, without first transcribing the speech to text has emerged as an interesting research problem. This work proposes such as system with an additional constraint of designing a system that has a small enough footprint to run on small micro-controllers and embedded systems with minimal latency. Given a streaming input speech signal, the proposed system can process it segment-by-segment without the need to have the entire stream at the moment of processing. The proposed system is evaluated on the publicly available Fluent Speech Commands dataset. Experiments show that the proposed system yields state-of-the-art performance with the advantage of low latency and a much smaller model when compared to other published works on the same task. 
\end{abstract}


\section{Introduction}

Nowadays, spoken language understanding (SLU) systems are crucial for daily life communication, where they can provide the crucial vocal user interface to home-controller devices and other appliances. The role of an SLU system is to convert a given speech signal to a structured representation, such as in the form of intent/slots classes, that could be interpreted  by a software and application to ultimately perform an action on the target device  \cite{tur2011spoken,bapna-etal-2017-sequential}. For example, a speech signal like `set an alarm for 5 p.m.' might have the following representation \{intent: `SET\_ALARM', data\_time: `5 P.M.'\}.

In classical SLU systems, an automatic speech recognition (ASR) model is first used to transcribe speech signals to a word string, followed by a natural language understanding (NLU) model that classifies this word string to the target intent representation. While this approach works well in various scenarios, there are a number of problems. For instance, as mentioned in \cite{8461718, ghannay:hal-01987740}, the two models, ASR and NLU, are often trained independently and not jointly optimized. Most ASR systems in themselves consists of a number of dis-jointly trained components. These issues can affect the overall performance of the SLU systems. Furthermore, these models often have high data and computational requirements, limiting their applicability to a handful of use-cases and languages. 

The number of recent end-to-end and ASR-free SLU systems show promising results \cite{haghani2018audio, lugosch2019speech, 8268987,Serdyuk2018TowardsES}. These SLU systems map a speech signal directly to the speaker's intent without explicitly recognizing the corresponding text. As with the conventional ASR+NLU systems, the end to end SLU systems are often computationally demanding. For example, the SLU systems presented in \cite{haghani2018audio, lugosch2019speech, 8268987,Serdyuk2018TowardsES} give good performance in terms of the recognition accuracy, however, they are a combination of several neural networks with each having hundreds of millions of parameters. Running such systems in real-time on low-power devices is not feasible.

In this work, we present an efficient and compact end-to-end SLU system. The proposed system is targeted at low-footprint devices, where the entire speech data is processed on the device without the need to send any data to a cloud server. By processing the entire speech data on the device, such a system provides increased privacy for the user. Furthermore, such a system can enable voice user interfaces for a number of use-cases and applications that would not have been possible in an always-connected scenario.  There are a number of technical contributions in this work that provide the aforementioned advantages. These are summarized below. 

\begin{itemize}
    \item In order to have the low latency property, the proposed model process a given input speech signal segment-by-segment. Here, the processing of one segment is done while receiving the upcoming one. The processing begins even if the speech signal is not entirely received. 
    
    \item The proposed model is built using convolutional layers, which are less computationally expensive than recurrent layers \cite{gao2019edgedrnn, zhang2015characterlevel,bai2018empirical}.
    
    \item The proposed model can process speech signals of variable duration without requiring any padding or cutting of the incoming speech.
 
\end{itemize}

\section{Related work}

In this section, we present the current state-of-the-art of the end-to-end SLU systems. 

In \cite{8461718}, the proposed SLU system is a speech-to-intent approach tested for semantic classification in dialog systems. In this system, the given speech signals are mapped directly to semantic meaning. This has several advantages. First, richer information than words can be extracted from speech. Second, the ability to extract semantic meaning from mixed language speech. The model proposed in \cite{8461718}, is composed of an acoustic model pre-trained with CTC loss to predict graphemes, and a semantic model pre-trained with the outputs of the acoustic model to predict intents. Once the two models are pre-trained, a full pipeline training (i.e., a fine tuning) is done on the entire architecture.  In \cite{lugosch2019speech}, a similar framework to \cite{8461718} is proposed, which is composed of three models. The three models are pre-trained respectively with phonemes, words, and intents. Then, they are fine tuned by training them together.

As opposed to \cite{8461718, lugosch2019speech}, in \cite{Serdyuk2018TowardsES} no pre-training is used, which is similar to our model. Here in \cite{Serdyuk2018TowardsES}, an encoder-decoder framework is proposed. The encoder is a multi-layer bidirectional recurrent layers. The decoder maps the output of the encoder to its corresponding intent class. To reduce the computational time, Serdyuk et al. used a sub-sampling for the Hidden activations along the time domain \cite{bahdanau2014neural, 44926}.  

Recently,  Haghaniet et al. \cite{haghani2018audio} proposed and compared four different SLU encoder-decoder based approaches, which are augmented by the attention mechanism \cite{bahdanau2014neural}. In all these proposed approaches, the mapping of speech signals to intents is formulated as a sequence-to-sequence problem \cite{SutskeverVL14, ChoMGBSB14}. The first approach maps audio features directly to their corresponding semantic sequence (domain/intent/arguments). In the second approach, the decoder outputs, for a given speech signal, not only its corresponding semantic sequence, but also its sequence of graphemes. The third has two decoder models: one outputs its semantic sequence and another outputs its sequence of graphemes. In the last one, called a multistage model, two stages are used where, in the first stage, the transcript is predicted, and in the second stage, the semantics are predicted. Here, the two stages are independently optimized and afterwards the whole system is fine tuned together. Haghaniet et al. \cite{haghani2018audio} conclude, after evaluations on real-world scenarios, that having an intermediate text representation and jointly optimizing the full system improves the overall accuracy of prediction.

In all these approaches \cite{8461718, lugosch2019speech, Serdyuk2018TowardsES, haghani2018audio}, the proposed network architectures are heavily based on recurrent layers, which may not be well suited for low-power devices \cite{Amoh2019AnOR}. These types of layers are slower and less computationally efficient than convolutional layers \cite{gao2019edgedrnn, zhang2015characterlevel,bai2018empirical}.

\section{Proposed Approach}
This section presents the proposed architecture along with problem formulation and the relevant implementation details.
\subsection{Problem formulation}

The speech-to-intent model proposed in this work is composed of a sequence of convolutional layers followed by a global max-pooling layer and few fully connected layers that output the intent class (see Table \ref{table_CNN_arch}). Here, the global max-pooling layer has crucial importance since it allows processing any given input speech signal segment-by-segment without the need to have it entirely at the moment of processing. In addition, this layer allows to process speech signals with different variable length, where no padding or concatenating is demanded.    

In Figure \ref{fig:tradvsseg}, we show the difference between processing a full speech signal versus segment-by-segment speech signal processing (i.e., the proposed scenario). As is shown, in the full-signal scenario, we wait till receiving the full speech signal then we process it (i.e., forwarded it through the model). However, in segment-by-segment signal processing scenario, each segment is forwarded separately through the convolutional layers and the global max-pooling layer. Then, all the outputs of all segments are stacked together and we again apply the max-pooling through them and we forward the resulting vector through the fully connected layers to classify the intent. Here, the advantage is that we can process one segment, while we are receiving the upcoming one. This results in less processing time, since a large part of the processing is done while receiving the speech signal.

For example, let $\{I_0,\ldots,I_T\}$ be the input acoustic features sequence for a given speech signal.  

\begin{itemize}
    \item In the full-signal processing scenario, the convolutional layers output a sequence $\{f_0,\ldots,f_N\}$ then the global max-pooling layer pool this sequence of outputs to one vector $R$, which is mapped by the fully connected layers to the intent class.
    
    \item In the segment-by-segment speech signal scenario (i.e., Algorithm \ref{algoTra}), the input acoustic features sequence can be viewed as two segments $\{I_0,\ldots,I_T\}= \{I_0,\ldots,I_s\} \cup \{I_{s+1},\ldots,I_T\}$, (i.e., in this example, there is no overlapping between segments). The convolutional layers output two sequence $\{f_0,\ldots,f_n\}$ and $\{f_0,\ldots,f_m\}$ for the two segments. Then, a global max-pooling layer pool the two sequences of outputs to two vector $R_1$ and $R_2$. In the last step, we apply again the global max-pooling for $\{R_1,R_2\}$ and outputs one vector $R$, which is mapped through the fully connected layers to the intent class. 
\end{itemize}

\begin{algorithm}[h]
\SetAlgoLined
	\KwIn{A stream input speech signals.}
	\KwOut{The intent class.}
	$T$=[]\;
    \While {waiting the upcoming segment ($S_{i+1}$)}
    {
    $R_i$=global-max-pooling(Conv-layers($S_i$))\;
    $T$=[$T$,$R_i$]\;
    }
    $R$=global-max-pooling(T)\;
    output=fully-connected($R$)\;
	\Return{output}
\caption{The segment-by-segment speech signal processing algorithm.}
\label{algoTra}
\end{algorithm}

The segment by segment processing is only used in the inference time. However, in the training time, the complete signal is processed at once. In the full-signal processing scenario, we need a total time of $T + \beta$ (seconds), where $T$ is the time to receive the speech signal and $\beta$ is the processing time. However, in the segment-by-segment speech signal scenario, the proposed approach needs less processing time. Since a part of the processing is done when receiving the speech signal. Here, the exact processing time depends on the two hyper-parameters: segment size and step size; these represent the size of the segment to process it each time and the size between the beginnings of two consecutive segments, respectively.  

\makeatletter
\newcommand*{\centerfloat}{%
  \parindent \z@
  \leftskip \z@ \@plus 1fil \@minus \textwidth
  \rightskip\leftskip
  \parfillskip \z@skip}
\makeatother

\begin{figure*}[ht!]
	\centerfloat
	\begin{footnotesize}

	\mbox{
		\shortstack{\fbox{\includegraphics[width=0.5\linewidth]{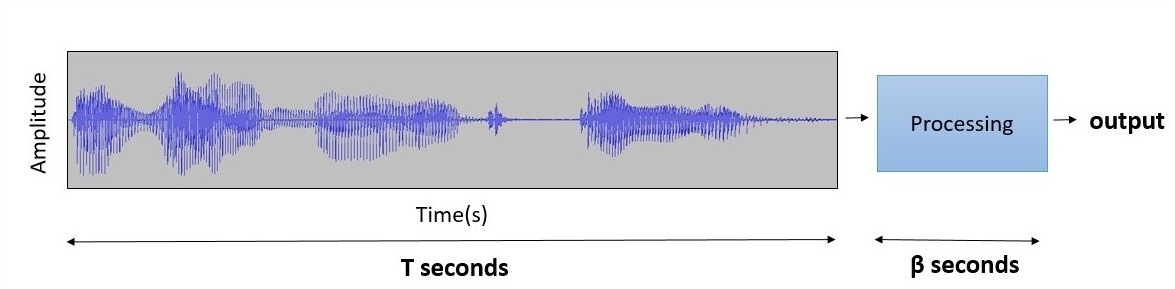}} \\[3mm] 
		    a) Full-signal processing scenario.
		}
		
		\hspace{5mm}
		
		\shortstack{\fbox{\includegraphics[width=0.5\linewidth]{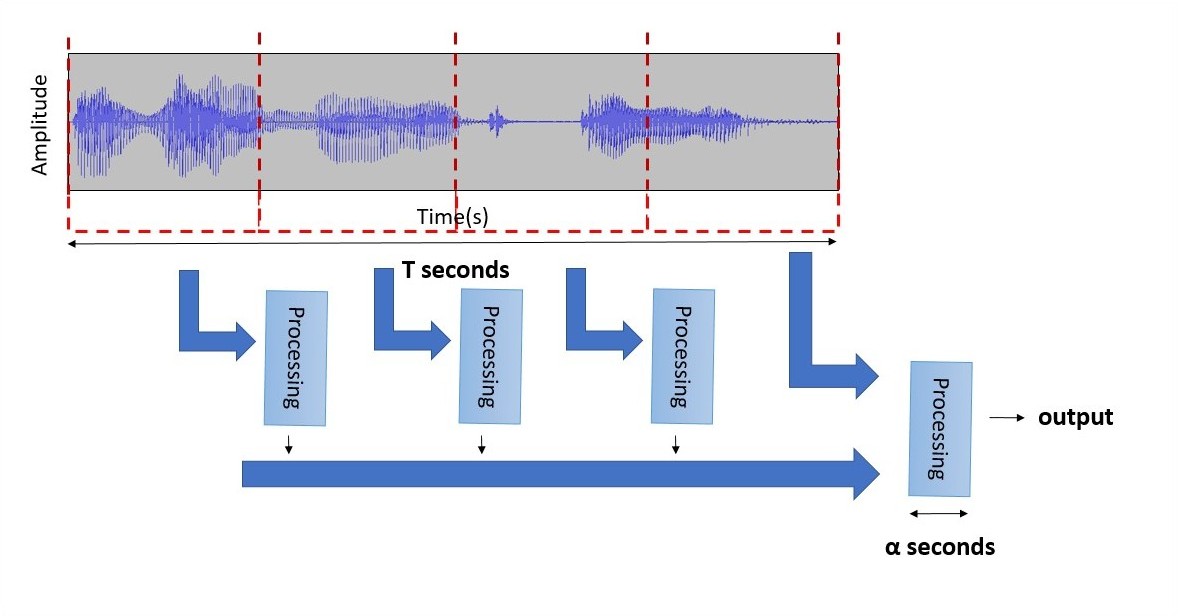}} \\[3mm] 
		    b) Segment-by-segment signal processing scenario (the proposed approach).
		}
    }
    \end{footnotesize}
	\caption{A comparison between traditional approaches that encode speech signal after receiving it fully and the proposed approach that encodes it segment-by-segment while receiving it. The first approach needs a total time \boldsymbol{$T + \beta$} (seconds) (T is the long of the speech signal), while the proposed approach needs a \boldsymbol{$T + \alpha$} (seconds). Here, \boldsymbol{$\beta>\alpha$} since a part of the processing is done when receiving the speech signal, which gives a low-latency SLU system.}
	\label{fig:tradvsseg}
	\vspace{-1em}
\end{figure*}

\subsection{The proposed network architecture}

Table \ref{table_CNN_arch} presents details about the proposed network architecture. Here, the global max-pooling layer is a critical component. It allows processing speech signals of variable duration segment-by-segment.  The proposed network is composed of 17 layers in total, with 8 convolutional layers, 4 max-pooling layers, 1 global max-pooling layer, and 4 fully-connected layers. All hidden layers use rectified linear units (ReLU) \cite{BengioCV13}. Moreover, as recommended in \cite{IoffeS15}, batch normalization is applied before each activation layer. The 8 convolutional layers with the 4 max-pooling layers can be viewed as 4 blocks. Where, each block is composed of one convolutional layer followed by a max-pooling layer of $(2\times1)$ kernel and a convolutional layer of $(1\times1)$ kernel. The  $(1\times1)$ convolutional layer is used mainly to reduce the number of features and to keep a small-footprint model.  

\begin{table}[ht]
\centering
\renewcommand{\arraystretch}{1}
\caption{The CNN architecture for speech segment of \\ size $100 \times 41$ (1 second).}
\label{table_CNN_arch}
\begin{footnotesize}
\begin{tabular}{lc}
\toprule
\textbf{Layer type} & \textbf{Output shape} \\
\midrule\midrule
4$\times$41, Conv2D, 128  & \fsz{128}{97}{1} \\
Max-pooling  &  \fsz{128}{48}{1} \\
1$\times$1, Conv2D, 64  & \fsz{64}{48}{1} \\
\midrule
4$\times$1, Conv2D, 128  & \fsz{128}{45}{1} \\
Max-pooling  &  \fsz{128}{22}{1} \\
1$\times$1, Conv2D, 64  & \fsz{64}{22}{1} \\
\midrule
4$\times$1, Conv2D, 128  & \fsz{128}{19}{1} \\
Max-pooling  &  \fsz{128}{9}{1} \\
1$\times$1, Conv2D, 64  & \fsz{64}{9}{1} \\
\midrule
4$\times$1, Conv2D, 256  & \fsz{256}{6}{1} \\
Max-pooling  &  \fsz{256}{3}{1} \\
1$\times$1, Conv2D, 256  & \fsz{256}{3}{1} \\
\midrule
\textbf{Global max pooling} & \textbf{256} \\
Fully-connected ~(ReLU units) & 256  \\
Fully-connected ~(ReLU units) & 196  \\
Fully-connected ~(ReLU units) & 128  \\
Fully-connected ~(Softmax) & 31$^*$\\
\midrule
\bottomrule
\multicolumn{2}{l}{$^*$ \emph{Number of intents in the Fluent  Speech  Commands dataset}.}
\vspace{-2em}
\end{tabular}
\end{footnotesize}
\end{table}

In Table \ref{table_CNN_arch}, an input of size $(100 \times 41)$ is used. This input represents one second of speech, where the acoustic features are extracted each 10ms for a speech signal frame sized of 25ms. The 41 elements are the 40 filters banks and the energy measure for the corresponding frame. As mentioned in \cite{VIIKKI1998133}, applying the cepstral mean and variance normalization (CMVN) on these 41 features improves speech recognition/classification performances. This normalization reduces the environmental changes and mismatch between the training and the testing conditions, where different background noises or different microphones can be presented.  In full-signal processing scenario, the CMVN is usually applied at the utterance level. However, in the proposed segment-by-segment speech signal scenario, utterance CMVN can not be applied because it requires having received the whole speech signal in the moment of processing.  Instead, we use globally computed mean and variance with all the training data \cite{Zeyer2016TowardsOW, 6707746}.

\section{Results}

 \begin{table}[ht]
  	\centering
 	\renewcommand{\arraystretch}{0.9}
 	\caption{Information  about  the  Fluent  Speech  Commands dataset.}
 	\label{table_dataset}
 		\begin{footnotesize}
 		\begin{tabular}{lccc}
 			\toprule
 			\textbf{Split} & \textbf{\# of speakers} & \textbf{\# of utterances} & \textbf{\# hours}   \\
 			\midrule\midrule
 			 Train & 77 & 23,132 & 14.7 \\
 			 Validation & 10 & 3,118 & 1.9 \\
 			 Test & 10 & 3,793 & 2.4 \\
 			\midrule
 			 \\
             Total & 97 & 30.043 & 19.0 \\
            \midrule
            \bottomrule
 		\end{tabular}
 		\end{footnotesize}
 \end{table}
 
This section summaries the experiments conducted in this work. First, we present the performance in terms of recognition accuracy of the proposed approach on the publicly available Fluent Speech Commands dataset \cite{lugosch2019speech}. Table \ref{table_dataset} summarizes some information about the dataset. Next, we evaluated the impact of different CMVN optimizations on the recognition accuracy. Finally, we evaluate the impact of various hyper-parameters on the proposed approach.

 \subsection{Comparison to state-of-the-art}

 \begin{table}[h]
  	\renewcommand{\arraystretch}{1.2}
  	\caption{ Error rate on the clean testing set comparing to state-of-the-art models using the Fluent Speech Commands dataset.}
  	\label{table_comp}
  	\centering
  	\begin{footnotesize}
  		\begin{tabular}{lcc}
  			\toprule
  			\textbf{Method} & \textbf{clean} & \textbf{Model size}  \\
  			\midrule \midrule
             Lugosch et al. (Full pipeline) \cite{lugosch2019speech} & 1.2\%  & 14.6M \\
             Lugosch et al. (No pre-training) \cite{lugosch2019speech} & 3.6\%  &  14.6M \\
             Poncelet et al. (Capsule net) \cite{poncelet2020multitask} & 1.9\%  &  - \\
             Palogiannidi et al. \cite{palogiannidi2019endtoend} & 1.38\%-5.83\%  &  - \\
            \midrule
             Proposed model & 2.18\%  &  \textbf{1.3M} \\
            \midrule
  			\bottomrule
  		\end{tabular}
  		\end{footnotesize}
\end{table}

Table \ref{table_comp} presents the results in terms of command recognition accuracy on the clean Fluent Speech Commands test set for the proposed approach along with other works from recent research. Lugosch et al. in \cite{lugosch2019speech} used a model composed of a hierarchy of sub-networks. The first sub-network is trained to recognize phonemes from acoustic features. The second one is trained to recognize words from phonemes. The last one is trained to recognize intents from words. In the full pipeline scenario, Lugosch et al. first pre-trained the three sub-networks separately then the entire model is fine-tuned. Our approach can be compared to the approach of Lugosch et al. \cite{lugosch2019speech}, where no pre-trained model is used. We can observer from the table that our approach has better performance (+1.42\%). In addition, our model is only $1.3 MB$ in size compared to $14.6 MB$ for Lugosch's model. The work of Poncelet el al. \cite{poncelet2020multitask} has a comparable performance to ours (1.9\% vs 2.18\%) similar to the work of Palogiannidi et al. \cite{palogiannidi2019endtoend}, which has performances between 1.38\% and 5.83\% depending on the number and the types of the recurrent layers. Our approach is more appropriate for low-power devices since it is built from only convolutional layers, where the models of \cite{lugosch2019speech,poncelet2020multitask,palogiannidi2019endtoend} includes many recurrent layers, which are slower compared to convolutional layers \cite{gao2019edgedrnn, zhang2015characterlevel,bai2018empirical}. 

\subsection{Comparison of using different CMVN normalization}

\begin{figure}[h]
	\centering
		\includegraphics[width=0.9\linewidth]{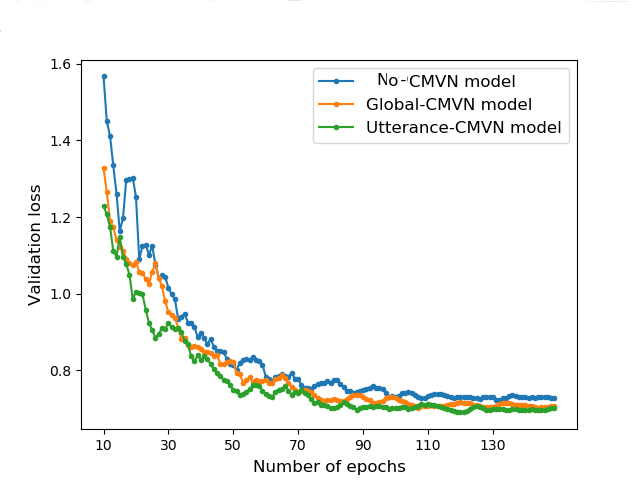}
	\caption{Validation loss for no CMVN, Global CMVN and Utterance CMVN models.}
	\label{CMVN}
\end{figure}

Figure \ref{CMVN} shows the validation losses for the three models trained respectively, with global CMVN, utterance CMVN and without any CMVN (No-CMVN). We can see that applying the global CMVN and utterance CMVN has almost the same performance. However, in the segment-by-segment processing scenario, we cannot apply utterance based CMVN, therefore, our only options are either applying the global CMVN or no CMVN at all. It is evident from the results in the Table \ref{table_cmvn} that applying the global CMVN leads to a higher performance. In the table, the column labeled `5dB' represents the testing case when the Fluent dataset is enhanced by augmenting with a mix noise types with a signal-to-noise ratio of 5dB. The `5dB + ff' scenario refers to the `5dB' set that is further augmented by a set of real Room Impulse Responses (RIRs). For the rest of this paper, all the experiments are performed with the global CMVN model. This is also true for the results presented for the proposed algorithm in Table \ref{table_comp}.
  
 \begin{table}[H]
  	\renewcommand{\arraystretch}{1.2}
  	\caption{ Error rate on the testing set for no CMVN, Global CMVN and Utterance CMVN models.}
  	\label{table_cmvn}
  	\centering
  	\begin{footnotesize}
  		\begin{tabular}{lccc}
  			\toprule
  			\textbf{Method} & \textbf{clean} & \textbf{5dB} & \textbf{5dB + ff} \\
  			\midrule \midrule
             no CMVN & 2.18\%  & 10.23\% & 19.11\% \\
             Global CMVN & \textbf{2.18\%}  & \textbf{9.96\%} & \textbf{18.98\%} \\
             \midrule
             Utterance CMVN & 2.48\%  & 9.1\% & 18.4\% \\
  			 \midrule
  			 \bottomrule
  		\end{tabular}
  		\end{footnotesize}
\end{table}

\subsection{Hyper-parameters analysis}

There are two main hyper-parameters in the segment-by-segment signal processing scenario. The first is the segment size, $S$, that represents the size of the audio segment to process. The second is the step size, $T$, that represents the amount which the window is moved between two consecutive segments. This means that each $T$ seconds, we process the last $S$ seconds till the ending of the speech signal. When, the step size is less than the segment size, this means that there is an overlapping between the segments. 

\begin{table}[H]
  	\renewcommand{\arraystretch}{1.2}
  	\caption{Mean Error rate on the (clean, 5dB and 5dB+ff) testing set for the segment-by-segment signal processing scenario using different hyper-parameters values.}
  	\label{table_hyper}
  	\centering
  	\resizebox{\columnwidth}{!}{
  	\begin{footnotesize}
  		\begin{tabular}{cccccc}
  			\toprule
  			\textbf{(Step $T\downarrow$ / Segment $S\rightarrow$)} & \textbf{1s} & \textbf{1.25s} & \textbf{1.5s} & \textbf{1.75s} & \textbf{2s}\\
  			\midrule \midrule
            \textbf{0.25s}  & \textbf{10.47\%}  & 10.61\% & 10.81\% & 10.81\% & 10.8\%\\
            \textbf{0.5s}   & 10.57\%  & 10.64\% & 10.54\% & \textbf{10.51\%} & 10.53\%\\
            \textbf{0.75s}  & 24.38\%  & 10.81\% & 10.47\% & \textbf{10.26\%} & 11.07\%\\
            \textbf{1s}     & 46.63\%  & 17.53\% & 11.43\% & 10.54\% & \textbf{10.53\%}\\
            \textbf{1.25s}  & 61.49\%  & 34.67\% & 20.85\% & \textbf{17.43\%} & 13.95\%\\
            \textbf{1.5s}   & 69.42\%  & 49.56\% & 38\%    & 15.93\% & \textbf{11.49\%}\\
  			\midrule
  			\bottomrule
  		\end{tabular}
  		\end{footnotesize}
    } 
\end{table}

Table \ref{table_hyper} shows that smaller step size, and hence the overlapping between segments, is crucial. A small step size means more information to encode, while information can be seen in different segments. This results in a better and richer representation. On the other hand, a higher step size means less computation. However, since a segment is processed while the system is receiving the next segment, having a high step size is not beneficial. Furthermore, we can observe that using a high segment size is not always beneficial. A small segment size also is not preferred since it can effect the overlapping. To conclude, the best hyper-parameters combination is the one that preserve a higher overlapping between segments. 

\begin{table}[H]
 	\caption{Mean Error rate on the (clean, 5dB and 5dB+ff) testing set for the segment-by-segment signal processing scenario versus the full-signal processing scenario.}
  	\centering
  	\resizebox{\columnwidth}{!}{
 	\label{table_segvsfull}
 		\begin{footnotesize}
 		\begin{tabular}{lcc}
 			\toprule
 			\textbf{hyper-parameters } & \textbf{Mean Error rate} & \textbf{\thead{Time of processing needed\\ after fully receiving the speech signal (\%)}}  \\
 			\midrule\midrule
 			 (Segment, Step)=(1.75s, 0.75s) & \textbf{10.26\%} & 43\% \\
 			 (Segment, Step)=(1s, 0.25s) & 10.47\% & \textbf{25\%} \\
 			 \midrule 
 			 Full signal speech & 10.37\% & 100\% \\
            \midrule
            \bottomrule
 		\end{tabular}
 		\end{footnotesize}
 }
 \end{table}

Table \ref{table_segvsfull} shows the difference in performances between the segment-by-segment signal processing scenario and the full signal processing scenario. We can see that the segment-by-segment processing not only reduces the time of processing but also improves the performance (10.26\% vs 10.37\%). Here, the time of processing needed after fully receiving the speech signal is represented as a percentage. In full signal processing scenario (Figure \ref{fig:tradvsseg}), it is represented by ($\beta$ seconds) equal to 100\%.  In segment-by-segment processing scenario with the hyper-parameters (Segment, Step)=(1.75s, 0.75s), the time of processing is 43\% of $\beta$. This means that after fully receiving the speech signal, we wait 2.3 times less with the segment-by-segment processing scenario than the full signal processing scenario. 

\section{Conclusions}

In this paper, we presented a speech-to-intent model for low-power devices. The proposed model has a global max-pooling layer that allows not only processing no fixed-length speech signals. But also, it allows processing any speech signal without the need to have it entirely at the moment of processing. The proposed processing scenario is done segment-by-segment, which means that the processing of one segment is done while we are receiving the upcoming segment. This not only reduces the response time but also improves the performances. 

\newpage  

\bibliographystyle{IEEEtran}

\bibliography{mybib}

\end{document}